\documentclass{article}

\usepackage[preprint]{neurips_2025}

\usepackage{latexsym}
\usepackage[T1]{fontenc}
\usepackage[utf8]{inputenc}
\usepackage{booktabs}
\usepackage{microtype}
\usepackage{makecell}
\usepackage{inconsolata}
\usepackage{xcolor}

\usepackage{arydshln}
\usepackage{amsmath}
\usepackage{graphics}
\usepackage{graphicx}
\usepackage{tikz}
\usepackage{wrapfig}
\usepackage{subcaption}
\usepackage{multirow}
\usepackage{pgfplots}
\usepackage{todonotes}
\usepackage{hyperref}
\usepackage{amsfonts}
\usepackage{nicefrac}
\usepackage{enumitem}
\usepackage{colortbl}
\usepackage{multicol}
\usepackage{listings}
\usepackage{tcolorbox}
\usepackage{url}

\pgfplotsset{compat=1.18}
\protected\edef\mathbb{\unexpanded\expandafter\expandafter\expandafter{\csname mathbb \endcsname}}

\usepackage[ethiop,main=english]{babel}
\newcommand\geez[1]{\selectlanguage{ethiop}{#1}\selectlanguage{english}}
\providecommand{\keywords}[1]{\textbf{\textit{Keywords:}} #1}

\title{Natural Language Processing for Tigrinya: \\Current State and Future Directions}

\author{
    Fitsum Gaim \hspace{6mm} Jong C. Park \\
    School of Computing \\
    Korea Advanced Institute of Science and Technology (KAIST) \\
    \texttt{\{fitsum.gaim, jongpark\}@kaist.ac.kr} \\
}
\date{\today}

\begin{document}

\maketitle

\begin{abstract}
\noindent
Despite being spoken by millions of people, Tigrinya remains severely underrepresented in Natural Language Processing (NLP) research. This work presents a comprehensive survey of NLP research for Tigrinya, analyzing over 50 studies from 2011 to 2025. We systematically review the current state of computational resources, models, and applications across fifteen downstream tasks, including morphological processing, part-of-speech tagging, named entity recognition, machine translation, question-answering, speech recognition, and synthesis. Our analysis reveals a clear trajectory from foundational, rule-based systems to modern neural architectures, with progress consistently driven by milestones in resource creation. We identify key challenges rooted in Tigrinya's morphological properties and resource scarcity, and highlight promising research directions, including morphology-aware modeling, cross-lingual transfer, and community-centered resource development. This work serves both as a reference for researchers and as a roadmap for advancing Tigrinya NLP. An anthology of surveyed studies and resources is publicly available.\footnote{The Tigrinya NLP Anthology: \url{https://github.com/fgaim/tigrinya-nlp-anthology}.}
\end{abstract}


\section{Introduction}

Natural Language Processing (NLP) has achieved remarkable progress in recent years, with breakthroughs in machine translation, question-answering, and language generation, transforming how humans interact with technology. However, these advances are largely confined to a small subset of the world's 7,000+ languages, creating a digital divide with significant societal implications \citep{hovy-spruit-2016-social, joshi-etal-2020-state, gaim-etal-2023-question}. Tigrinya (\geez{tgr~nA}; ISO 639-3: \texttt{tir}), spoken primarily in Eritrea and Ethiopia, exemplifies the challenges faced by most underrepresented languages in the digital age.

The lack of comprehensive surveys for specific low-resource languages creates barriers for new researchers and hinders systematic advancement. While recent reviews have partially covered related languages \citep{tonja-etal-2023-natural}, no prior work has presented a focused, comprehensive analysis of Tigrinya NLP research. This paper addresses that gap by charting the field's evolution from early rule-based morphological analyzers to modern neural language models, highlighting how community-led development of computational resources has been the primary catalyst for advancement.

We present a comprehensive survey of NLP research for Tigrinya, analyzing over 50 studies published between 2011 and 2025. The contributions of this work include:

\begin{itemize}
    \item A systematic review of Tigrinya NLP research across fifteen downstream tasks, analyzing the chronological development of methods and resources.
    \item An analysis of datasets, tools, and pre-trained models that have enabled progress.
    \item The identification of critical research gaps and recommendations for future work.
    \item A replicable methodology for surveying progress in other low-resource languages.
\end{itemize}

\section{Tigrinya Language: Characteristics and Computational Challenges}

\subsection{Linguistic Background}

Tigrinya is a Semitic language of the Afro-Asiatic family, serving as a national language in Eritrea and a regional language in Tigray, Ethiopia \citep{negash-2016-tigrinya-origin}. With an estimated 10 million native speakers worldwide \citep{eberhard-etal-2025-ethnologue}, it represents a significant yet computationally underserved language community. The language is part of a linguistically rich area known as the \textit{North East African Language Macro-Area}, characterized by shared features across different language families, which influences its structure and evolution \citep{zaborski-2011-language-subareas}.

\subsection{Writing System and Orthography}

\paragraph{The Ge'ez Script.} Tigrinya uses the Ge'ez script (\geez{fidale}, \textit{fid\"al}), an ancient writing system with a rich history in the Horn of Africa. The oldest known example of the Ge'ez script is found on the obelisk (Hawulti) at Matara in modern-day Eritrea, dated to the early 4th century CE \citep{ullendorff-1951-obelisk-of-matara}. The Ge'ez script is an \textit{abugida}, where each character represents a consonant-vowel syllable. These characters are formed by systematically modifying a base consonant grapheme to denote one of seven vowels. The Tigrinya syllabary comprises 32 base consonants, each with seven vowel forms (or orders), supplemented by five labialized consonant sets with five vowel forms each, yielding a total of 249 distinct characters~($32 \times 7 + 5 \times 5 = 249$). While the Ge'ez script's unique punctuation system is retained in modern Tigrinya, its native numerals have been supplanted by Western Arabic numerals.

The Ge'ez orthography presents distinct challenges for computational processing. First, unlike Latin-based scripts, Ge'ez lacks a letter case distinction (i.e., no capitalization), which complicates acronym detection and named entity recognition (NER). Second, the script does not explicitly mark gemination (consonant doubling), a phonemic feature in Tigrinya that distinguishes word meanings \citep{gasser-etal-2011-hornmorpho}. For instance, \textit{q\"ar\"ab\"a} (``he approached'') and \textit{q\"arr\"ab\"a} (``he offered'') are both written identically as \geez{qaraba}, i.e., they are \textit{heterophonic homographs}, also known as \textit{heteronyms}, words that share spelling but differ in pronunciation and meaning. This orthographic ambiguity must be resolved based on context, posing a significant challenge for natural language understanding (NLU) tasks such as morphological analysis, part-of-speech (POS) tagging, named entity recognition (NER), as well as for speech processing applications such as text-to-speech (TTS) synthesis.

\paragraph{Digital Orthography.} In the digital age, typing in Tigrinya has been facilitated by specialized input methods. Software programs such as GeezWord\footnote{GeezWord software keyboards by GeezSoft: \url{https://geezsoft.com}}, GeezIME\footnote{GeezIME software keyboards by GeezLab: \url{https://geezlab.com}}, and Keyman\footnote{Keyman Keyboards for Tigrinya: \url{https://keyman.com/keyboards/h/tigrigna}} allow users to type Ge'ez script-based languages such as Tigrinya, Amharic, and Tigre using standard QWERTY keyboards. These tools employ variations of phonetic systems that map combinations of ASCII symbols to the syllabic Ge'ez characters \citep{firdyiwek-yaqob-2000-system}. Understanding the conventions set by these input methods is crucial for NLP tasks such as text normalization, spell checking, and data preprocessing.

\subsection{Morphological Complexity}

Tigrinya exhibits both templatic (root-and-pattern) and agglutinative morphological processes, creating a sophisticated system that is challenging for computational models.

\paragraph{Templatic Morphology.} Words are formed by inserting a consonantal root, typically of three consonants, into various vowel patterns. This non-concatenative process allows generation of numerous surface forms from a single root. Some examples derived from the root \textit{s-b-r} (\geez{sbr}) related to the concept of ``breaking'' include:
\begin{itemize}
    \item \textit{s\"ab\"ar\"a} (\geez{sabara}) - ``he broke'' (perfective)
    \item \textit{yis\"abbir} (\geez{yesabere}) - ``he breaks'' (imperfective)
    \item \textit{sibur} (\geez{sebure}) - ``broken'' (adjective)
    \item \textit{m\"asb\"ar} (\geez{masebare}) - ``place of breaking'' (instrumental noun)
\end{itemize}

\paragraph{Agglutinative Features.} Tigrinya also attaches prefixes, suffixes, and infixes to word stems to modify meaning. These affixes are generally separable and carry consistent semantic categories across different roots. Some examples include:
\begin{itemize}
    \item The prefix \geez{ta-} (\textit{t\"a-}) forms passive or reflexive verbs: \textit{s\"ab\"ar\"a} (``he broke'') $\to$ \textit{t\"as\"ab\"ar\"a} (\geez{tasabara}, ``it was broken'').
    \item Suffixes like \geez{-kA} (\textit{-ka}) and \geez{-ki} (\textit{-ki}) mark possession: \textit{bet} (\geez{bEte}, ``house'') $\to$ \textit{betka} (\geez{bEtekA}, ``your house,'' masc.) and \textit{betki} (\geez{bEteki}, ``your house,'' fem.).
\end{itemize}

\paragraph{Grammatical Categories and Computational Impact.} This morphological richness allows a single inflected word to express numerous grammatical categories such as gender, number, and tense. This process results in rapid vocabulary growth and severe data sparsity due to the high type-token ratios observed in naturally available text collections~\citep{gaim-etal-2021-tlmd}. These characteristics pose significant challenges for statistical and neural NLP methods that rely on word frequencies and fixed vocabularies.

\section{Current State of Tigrinya NLP Research}

\begin{figure}[!t]
\centering
\includegraphics[width=\columnwidth]{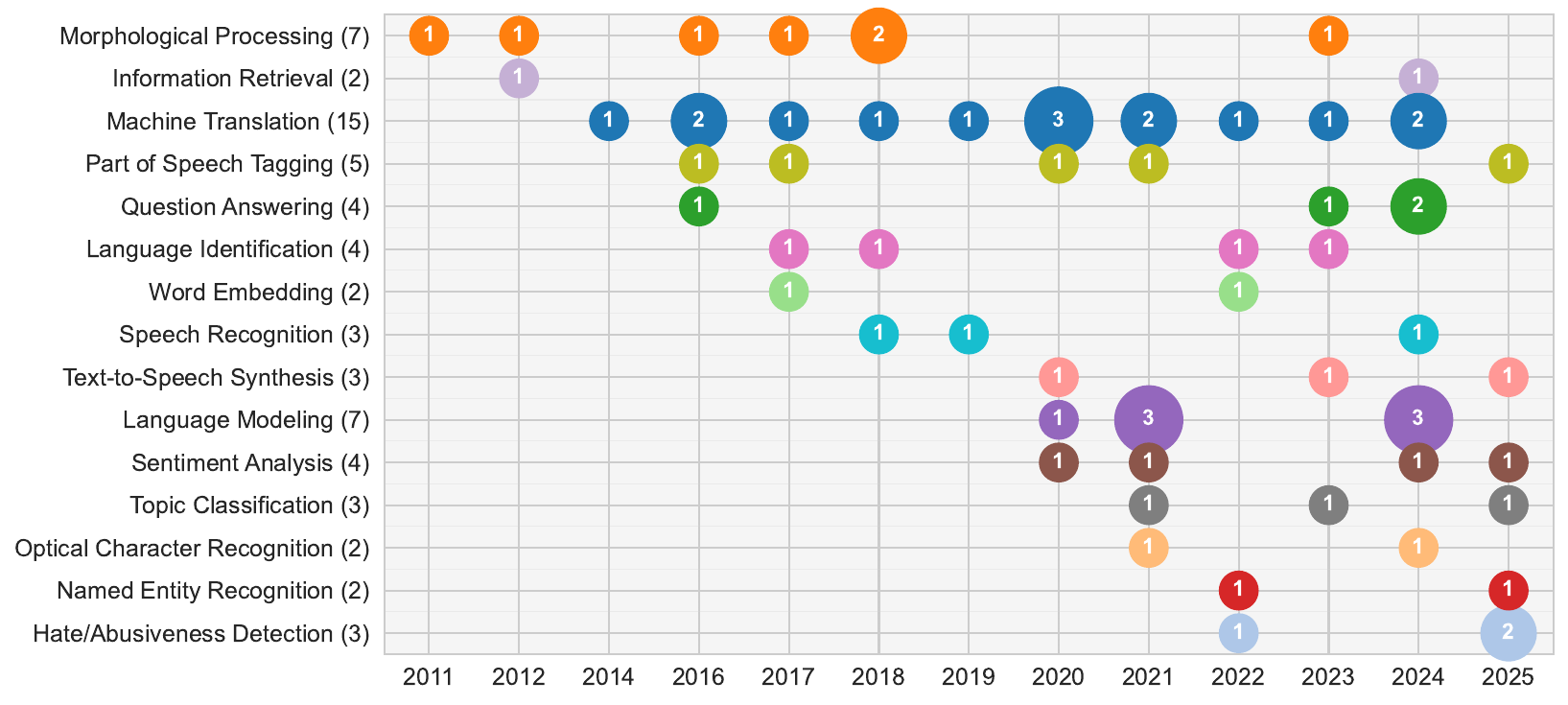}
\caption{Timeline and Distribution of Tigrinya NLP Research by Task Area (2011-2025). The number of publications in a year is indicated by the bubble size.}
\label{figure:nlp_timeline_distribution}
\end{figure}

Our analysis identified a clear evolution from foundational, rule-based systems to the adoption of modern neural architectures, driven by key resource creation milestones. Figure~\ref{figure:nlp_timeline_distribution} shows the chronological progression and distribution of publications. The research landscape is further detailed in Table~\ref{table:methodological_contributions}, which summarizes the key methodological contributions by task, and Table~\ref{table:tigrinya_nlp_assets}, which lists major publicly available datasets.

\subsection{Morphological Processing}
Morphological analysis has been a foundational research area, reflecting the central role of Tigrinya's complex linguistic structure in computational processing. Early work includes rule-based stemmers for information retrieval \citep{osman-mikami-2012-stemming} and comprehensive finite-state transducer systems like HornMorpho for analysis and generation \citep{gasser-etal-2011-hornmorpho}. Other approaches have explored parser combinators \citep{littell-etal-2018-parser}, neural methods using LSTMs for morpheme boundary detection \citep{tedla-etal-2018-lstm-segmentation}, along with a new morphological segmentation dataset \citep{gebremeskel-etal-2023-hybrid-morphological-analyzer} to further support this task.

\subsection{Machine Translation (MT)}
The largest share of studies ($\sim$30\%) have focused on machine translation, driven by its potential to improve information accessibility for the Tigrinya-speaking community. Early work focused on Statistical Machine Translation (SMT), applying morphological segmentation to mitigate data sparsity \citep{tedla-etal-2016-segmentation-mt, gaim-2017-morphology-mt}, exploring factored models \citep{tsegaye-2014-factored-mt}, and creating parallel corpora \citep{abate-etal-2018-parallel-corpora, abate-etal-2019-english-ethiopian-smt}. Hybrid approaches combining SMT with syntactic reordering rules achieved BLEU scores up to 32.64 \citep{berihu-2020-mt}. More recent research has shifted to Neural Machine Translation (NMT), investigating the Transformer architecture \citep{isayas-2021-nmt}, transfer learning from other Ge'ez script languages \citep{oktem-etal-2020-mt-transfer}, and data augmentation through back-translation \citep{kidane-etal-2021-augmentation, hadgu-etal-2022-lesan}. Despite this progress, performance is constrained by limited parallel corpora, with most studies using fewer than 20K sentence pairs. Error analyses confirm that mistranslation and omission remain frequent issues \citep{abdelkadir-etal-2023-mt-error}. Concurrent with these specific efforts, Tigrinya has been included in the NLLB project~\citep{meta-etal-2024-nllb}, which provides both a large training corpus of 1.4M sentence pairs and the \textsc{Flores-200} evaluation benchmark with 3K samples, leveraging scale to improve zero-shot translation. Community-driven evaluation of such global models remains a necessity to understand their performance and limitations.

\subsection{Part-of-Speech (POS) Tagging}
Initial POS tagging research was enabled by the Nagaoka Tigrinya Corpus \citep{tedla-etal-2016-pos-tagging}, a manually annotated dataset of 4,656 sentences. This resource facilitated the application of CRF and SVM models, and subsequent work analyzed the effectiveness of word embeddings for the task \citep{tedla-etal-2017-analyzing-we}. Deep learning models, particularly BiLSTMs, were later shown to be effective \citep{tesfagergish-2020-pos-dnn}. The current state-of-the-art was established by fine-tuning transformer-based monolingual pre-trained language models, which achieved 95.49\% accuracy \citep{gaim-etal-2021-tiplms}.

\subsection{Named Entity Recognition (NER)}
Research on Tigrinya NER has evolved from investigations with small annotated datasets \citep{yohannes-etal-2022-ner} to the development of large-scale, comprehensive resources such as the TiNC24~\citep{berhane-etal-2025-tinc24}. The TiNC24 dataset, with over 200K annotated entities, has enabled the training of robust sequence labeling models, achieving an F1 score of 90.18\% by fine-tuning pre-trained Tigrinya language models.

\subsection{Text Classification: Topic Classification, Sentiment Analysis, Hate/Abuse Detection}
Research in text classification has addressed various downstream tasks such as topic classification, sentiment analysis, hate speech detection, and abusive language detection. Early work focused on news categorization using CNNs \citep{fesseha-etal-2021-text-classification}. More recent efforts have targeted user-generated content, including sentiment analysis of social media comments via transfer learning \citep{tela-etal-2024-transferring-monolingual-model} and hate speech detection on Facebook \citep{bahre-2022-hate}. A multi-task benchmark for abusive language detection in Tigrinya (TiALD) was recently introduced with joint annotations for abusiveness, sentiment, and topic classification of YouTube comments \citep{gaim-etal-2025-tiald}. Similarly, a multilingual collection for hate speech detection in 15 African languages (AfriHate) includes a Tigrinya subset based on Twitter/X data \citep{muhammad-etal-2025-afrihate}.

\subsection{Question-Answering (QA)}
Tigrinya QA has seen significant progress in recent years. Starting from early explorations of factoid QA using SMT techniques \citep{amare-2016-question}, the area advanced substantially with the release of several dedicated resources: TiQuAD, a large-scale native-annotated reading comprehension benchmark \citep{gaim-etal-2023-question}; TIGQA, an expert-annotated educational dataset \citep{teklehaymanot-etal-2024-tigqa}; and Belebele, a multilingual multiple-choice QA dataset with a Tigrinya slice \citep{bandarkar-etal-2024-belebele}. These resources have facilitated experimentation with cross-lingual and multilingual approaches, with the best models achieving up to 85.12\% F1 score on the TiQuAD test set.

\subsection{Language Modeling}
The development of monolingual corpora, such as the TLMD~\citep{gaim-etal-2021-tlmd}, has been crucial. This resource enabled the pre-training of the first Tigrinya-specific language models (TiRoBERTa, TiBERT, TiELECTRA) \citep{gaim-etal-2021-tiplms}, which were shown to significantly outperform multilingual alternatives on various downstream NLU tasks. In parallel, another line of work explored multilingual encoders, notably AfriBERTa~\citep{ogueji-etal-2021-small}, trained on a collection of 11 African languages including Tigrinya, achieving promising results in several low-resource languages. More recently, the field has shifted toward massive multilingual generative models. The Aya Collection~\citep{singh-etal-2024-aya} is a multilingual instruction tuning dataset covering 114 languages including Tigrinya, enabling the development of chat-focused language models.

\subsection{Automatic Speech Recognition (ASR)}

Early efforts in Automatic Speech Recognition for Tigrinya focused on corpus design and the application of deep neural networks \citep{abera-h-mariam-2018-design, abera-h-mariam-2019-speech}. Further progress was made with the development of an end-to-end hybrid CTC-Attention model \citep{ghebregiorgis-etal-2024-tigrinya-asr}, which constructed a new 30-hour speech corpus and achieved a Word Error Rate (WER) of 36.01\%, demonstrating the viability of modern ASR architectures for the language.

\subsection{Text-to-Speech (TTS)}

Initial work in Text-to-Speech explored concatenative synthesis approaches \citep{keletay-worku-2020-tts}. More recently, end-to-end neural TTS has been explored using the Tacotron architecture, trained on a new 17.3-hour single-speaker corpus, achieving a Mean Opinion Score (MOS) of 3.41 on human testing \citep{mihreteab-etal-2025-end-to-end-tts}. A notable development is the inclusion of Tigrinya in Meta's Massively Multilingual Speech (MMS) project \citep{pratap-etal-2023-mms-tts-tir}, which provides an open-source adversarial learning based TTS model.

\subsection{Information Retrieval (IR)}

Information Retrieval remains one of the underexplored research areas in Tigrinya NLP. Early work explored word stemming methods for Lucene-based lexical retrieval systems \citep{osman-mikami-2012-stemming}. More recently, \citet{gaim-2024-semantic-search} made available monolingual bi-encoder models for semantic representation of Tigrinya text that were trained on question-answering datasets. These models can be used for semantic search and retrieval-augmented generation (RAG) applications. However, a research gap remains in establishing standard benchmarks and developing effective methods for reranking to advance information retrieval for Tigrinya.

\subsection{Language and Dialect Identification}

Research in Language and Dialect Identification has included distinguishing between Tigrinya and typologically related languages that share the Ge'ez script \citep{asfaw-2018-lid, gaim-etal-2022-geezswitch}, identifying regional dialects of Tigrinya \citep{gedamu-2023-tigrinya-dialect}, and investigating mutual intelligibility with the Amharic language \citep{feleke-2017-similarity}.

\subsection{Optical Character Recognition (OCR)}

Optical character recognition for Tigrinya presents unique challenges due to the visual similarity between certain characters. The GLOCR dataset \citep{gaim-2021-glocr} provided a foundational resource, containing over 710K image-text pairs derived from diverse sources including news articles, books, and synthetic word trigrams. Building on this resource, \citet{hailu-etal-2024-tigrinya-ocr} applied CRNN-based models for text recognition, achieving high accuracy on printed Tigrinya text. However, handwritten text recognition and document layout analysis remain underexplored areas.

\begin{table*}[pt!]
\centering
\small
\caption{Methodological Contributions in Tigrinya NLP Research by Task Area}
\label{table:methodological_contributions}
\begin{tabular}{@{}p{3.3cm}p{10.2cm}@{}}
\toprule
\textbf{Task Area} & \textbf{Methodological Contributions \& Milestones} \\
\midrule
Morphological Processing & Rule-based stemming; Finite-State Transducers (FST); Parser combinators; Neural boundary detection (LSTM); Hybrid rule-neural analyzers. \textit{\citep{gasser-etal-2011-hornmorpho, osman-mikami-2012-stemming, littell-etal-2018-parser, tedla-etal-2018-lstm-segmentation, gebremeskel-etal-2023-hybrid-morphological-analyzer}} \\
\addlinespace
Machine Translation & Factored SMT; Hybrid SMT-Morphology systems; Neural MT; Transfer learning from related languages; Data augmentation (back-translation). \textit{\citep{tsegaye-2014-factored-mt, tedla-etal-2016-segmentation-mt, gaim-2017-morphology-mt, berihu-2020-mt, isayas-2021-nmt, kidane-etal-2021-augmentation, hadgu-etal-2022-lesan}} \\
\addlinespace
Part-of-Speech Tagging & Morphological segmentation; CRF, SVM \& Word Embeddings; Bi-LSTM \& CNN models; Fine-tuning of Pre-trained Language Models (PLMs). \textit{\citep{tedla-etal-2016-pos-tagging, tedla-etal-2017-analyzing-we, tesfagergish-2020-pos-dnn, gaim-etal-2021-tiplms}} \\
\addlinespace
Text Classification & Sentiment analysis for social media text; Applying CNNs for news classification; Efficient data sampling method abusive language detection; Multi-task learning for abusive, sentiment, and topic classification. \textit{\citep{tela-etal-2024-transferring-monolingual-model, fesseha-etal-2021-text-classification, bahre-2022-hate, gaim-etal-2025-tiald}} \\
\addlinespace
Language/Dialect ID & Mutual intelligibility studies; Dialect detection; Benchmark for typologically related languages; N-gram models; Fine-tuning of PLMs for language identification. \textit{\citep{feleke-2017-similarity, asfaw-2018-lid, gaim-etal-2022-geezswitch, gedamu-2023-tigrinya-dialect}} \\
\addlinespace
Named Entity Recognition & Small and large-scale annotated datasets; Joint POS tagging; Fine-tuning of PLMs for sequence labeling. \textit{\citep{yohannes-etal-2022-ner, berhane-etal-2025-tinc24}} \\
\addlinespace
Question-Answering & Factoid QA via SMT; Large-scale native extractive QA datasets; Expert-annotated educational datasets; Cross-lingual transfer. \textit{\citep{amare-2016-question, gaim-etal-2023-question, teklehaymanot-etal-2024-tigqa, bandarkar-etal-2024-belebele}} \\
\addlinespace
Language Modeling & Construction of large monolingual corpora; Pre-training of monolingual Transformer-based models; Multilingual modeling for low-resource African languages. \textit{\citep{gaim-etal-2021-tlmd, gaim-etal-2021-tiplms, ogueji-etal-2021-small}} \\
\addlinespace
Optical Character \& \newline Text Recognition & Development of large-scale labeled text-image dataset for Ge'ez script; Application of CRNN-based models to text recognition. \textit{\citep{gaim-2021-glocr, hailu-etal-2024-tigrinya-ocr}} \\
\addlinespace
Speech Recognition & Design and construction of speech corpus; Application of deep neural networks to speech recognition; End-to-end hybrid CTC-Attention ASR models. \textit{\citep{abera-h-mariam-2018-design, abera-h-mariam-2019-speech, ghebregiorgis-etal-2024-tigrinya-asr}} \\
\addlinespace
Text-to-Speech & Concatenative synthesis approaches; End-to-end neural TTS using Tacotron architecture; Massively Multilingual Speech (MMS) project with adversarial learning-based TTS. \textit{\citep{keletay-worku-2020-tts, mihreteab-etal-2025-end-to-end-tts, pratap-etal-2023-mms-tts-tir}} \\
\addlinespace
Information Retrieval & Word stemming for lexical retrieval; Monolingual bi-encoder models for semantic search and text representation. \textit{\citep{osman-mikami-2012-stemming, gaim-2024-semantic-search}} \\
\bottomrule
\end{tabular}
\end{table*}

\subsection{Other Downstream Tasks}

Studies across several task areas have employed word embeddings to address the challenges of morphological complexity and data scarcity in Tigrinya \citep{tedla-etal-2017-analyzing-we, fesseha-etal-2021-text-classification, tesfagergish-2020-pos-dnn, gaim-etal-2025-tiald}. To facilitate the systematic evaluation of such word embedding models, a dedicated analogy test set was developed for Tigrinya \citep{gaim-etal-2022-analogy-test} by translating and refining an English dataset.
In a related direction, initial evaluations of gender bias in Machine Translation systems found that 80\% of sentences exhibited gender bias \citep{sewunetie-etal-2024-gender-bias}, an expected challenge given Tigrinya's grammatical gender system that often defaults to masculine when not specified.

\subsection{Datasets and Resources}

A substantial portion of the reviewed literature is characterized by the concurrent development of linguistic resources such as datasets, benchmarks, and models. These contributions have been instrumental in advancing the Tigrinya NLP field. Table~\ref{table:tigrinya_nlp_assets} summarizes some of the major resources, highlighting the community's focus on building a robust data infrastructure for Tigrinya.

\begin{table*}[pt!]
\centering
\small
\caption{Overview of Publicly Available Resources for Tigrinya NLP Research: Showing the diversity of tasks addressed and the scale of community-contributed linguistic assets.}
\label{table:tigrinya_nlp_assets}
\resizebox{\textwidth}{!}{
\begin{tabular}{@{}llll@{}}
\toprule
\textbf{Resource} & \textbf{Task Area} & \textbf{Type} & \textbf{Size} \\
\midrule
NTC \citep{tedla-etal-2016-pos-tagging} & POS Tagging & Dataset & 4.6K sents, 72K tokens \\
\addlinespace
Sentiment \citep{tela-etal-2024-transferring-monolingual-model} & Sentiment Analysis & Dataset & 50K samples \\
\addlinespace
TLMD \citep{gaim-etal-2021-tlmd} & Language Modeling & Dataset & 2M sents, 40M tokens \\
\addlinespace
TiQuAD \citep{gaim-etal-2023-question} & Question Answering & Dataset & 10.6K QA pairs \\
\addlinespace
TIGQA \citep{teklehaymanot-etal-2024-tigqa} & Question Answering & Dataset & 2.6K QA pairs \\
\addlinespace
Belebele \citep{bandarkar-etal-2024-belebele} & Question Answering & Dataset & 900 QA pairs \\
\addlinespace
NER \citep{yohannes-etal-2022-ner} & Named Entity Recognition & Dataset & 3.6K sents, 69K entities \\
\addlinespace
TiNC24 \citep{berhane-etal-2025-tinc24} & Named Entity Recognition & Dataset & 13K sents, 200K entities \\
\addlinespace
Analogy Test \citep{gaim-etal-2022-analogy-test} & Word Embedding Evaluation & Dataset & 18.5K entries \\
\addlinespace
GeezSwitch \citep{gaim-etal-2022-geezswitch} & Language Identification & Dataset & 15K samples \\
\addlinespace
MasakhaNEWS \citep{adelani-etal-2023-masakhanews} & Topic Classification & Dataset & 3K articles \\
\addlinespace
TiALD \citep{gaim-etal-2025-tiald} & Abusive Language Detection & Dataset & 13.7K comments \\
\addlinespace
AfriHate \citep{muhammad-etal-2025-afrihate} & Hate Speech Detection & Dataset & 5K tweets (tir subset) \\
\addlinespace
\textsc{Flores-200} \citep{meta-etal-2024-nllb} & Machine Translation & Dataset & 3K sents \\
\addlinespace
NLLB \citep{meta-etal-2024-nllb} & Machine Translation & Dataset & 1.4M sents \\
\addlinespace
ASR Data \citep{abera-h-mariam-2018-design} & Speech Recognition & Dataset & 10K utterances \\
\addlinespace
GLOCR \citep{gaim-2021-glocr} & OCR & Dataset & 710K text-image pairs \\
\addlinespace
TiPLMs \citep{gaim-etal-2021-tiplms} & Language Modeling & Model & TiRoBERTa (125M); \\
 &  & & TiBERT (110M); \\
 &  & & TiELECTRA (14M) \\
\addlinespace
AfriBERTa \citep{ogueji-etal-2021-small} & Language Modeling & Model & AfriBERTa-large (126M); \\
 &  & & base (111M); small (97M) \\
\addlinespace
Tigrinya BiEncoders \citep{gaim-2024-semantic-search} & Information Retrieval & Model & TiRoBERTa-bi-encoder (125M); \\
 &  & & TiELECTRA-bi-encoder (14M) \\
\addlinespace
MMS-TTS-Tir \citep{pratap-etal-2023-mms-tts-tir} & Text-to-Speech & Model & VITS-based TTS (36M) \\
\bottomrule
\end{tabular}
}
\end{table*}

\section{Challenges and Future Directions}

Our analysis of the Tigrinya NLP landscape reveals a set of interconnected challenges and a clear roadmap for future progress. We consolidate our findings on challenges, opportunities, and research gaps into a unified discussion below.

\subsection{Challenges and Methodological Opportunities}

The primary obstacles in Tigrinya NLP are characteristic of many low-resource languages \citep{joshi-etal-2020-state}, but are amplified by specific linguistic features.

\paragraph{Data Scarcity and Morphological Complexity.} The most significant barrier is the combination of these two factors. The lack of large-scale, annotated datasets is severely exacerbated by Tigrinya's complex morphology, which leads to high out-of-vocabulary (OOV) rates and extreme data sparsity. This combination challenges standard tokenization and modeling techniques, making them less effective without massive datasets. This points toward methodological opportunities that are becoming central to low-resource NLP research.
\textbf{Data-efficient learning} offers a key strategy for leveraging multilingual models through transfer learning \citep{ruder-2019-neural} and cross-lingual methods to bootstrap performance. Similarly, synthetic data such as machine-translated silver-standard datasets can effectively augment natively annotated corpora.
Another important opportunity lies in designing \textbf{morphology-aware architectures}. Future work could investigate hybrid tokenization schemes (e.g., BPE~\citep{sennrich-etal-2016-neural} combined with morphological segmentation) or character-level models that are inherently better suited to Tigrinya's non-concatenative structure.

\paragraph{Limited Standardized Resources.} The field suffers from a scarcity of robust, open-source preprocessing tools and a lack of multi-domain evaluation benchmarks, making direct model comparison difficult. A key opportunity lies in \textbf{community-centered resource development}. Engaging with local Tigrinya-speaking institutions and the diaspora through participatory design can not only scale data collection but also ensure cultural relevance and help mitigate bias from the ground up \citep{bird-2020-decolonising}.

\paragraph{Societal Bias Amplification.} As with many languages, models trained on historical or web-crawled data risk amplifying societal biases. Initial studies have confirmed significant gender bias in MT systems \citep{sewunetie-etal-2024-gender-bias}. This frames a critical future direction: moving beyond performance metrics to responsible and fair NLP. This includes not just auditing for gender bias, but also for dialectal bias (given the known dialectal variations \citep{gedamu-2023-tigrinya-dialect}) and potential political or social biases inherent in news-centric corpora.

\subsection{Research Gaps and Recommendations}

Building on these challenges and opportunities, our analysis reveals several underexplored areas where future research could have a high impact. These gaps include a scarcity of dedicated work in \textbf{conversational} and \textbf{multimodal} applications. For instance, locally developed and culturally grounded conversational agents remain unexplored. Furthermore, research in \textbf{speech processing} is limited to initial prototypes of speech recognition and text-to-speech synthesis emerging recently through dedicated corpora and neural models. Finally, existing resources are heavily concentrated on news text, leaving a need for \textbf{domain adaptation} to critical areas like healthcare and law, and work on \textbf{bias and fairness} remains in early stages.

To address these gaps, we recommend a strategy focusing on technical priorities and community engagement. The community should prioritize \textbf{developing foundational resources and tools}, such as expanding parallel corpora, creating multi-domain evaluation benchmarks, and standardizing open-source preprocessing libraries. Concurrently, researchers could investigate \textbf{language-adaptive methods}, including morphology-aware neural architectures and conducting comprehensive audits to mitigate gender, cultural, and other societal biases. Lasting progress, however, depends on \textbf{establishing community partnerships} with local universities and cultural institutions to build local research capacity and prioritize applications that address direct community needs, e.g., educational tools and healthcare information access. Finally, all created resources, such as datasets, models, and tools, must be \textbf{open and accessible}, with meticulous documentation to empower the next wave of researchers and developers.

\section{Survey Methodology}

\subsection{Literature Search}
We conducted a comprehensive search across multiple databases including Google Scholar, Semantic Scholar, ACL Anthology, JSTOR, IEEE Xplore, and ACM Digital Library. Search terms included combinations of ``Tigrinya'', ``Natural Language Processing'', ``Computational Linguistics'', ``Machine Learning'', and specific task names. We also included published linguistic resources and grey literature such as theses and technical reports to ensure comprehensive coverage.

\subsection{Inclusion and Exclusion Criteria}
For a study to be included in this survey, we required that its primary focus be on Natural Language Processing or Computational Linguistics, with a substantial treatment of the Tigrinya language. Eligible works were further limited to original research articles and systematic reviews to ensure a focus on novel contributions and rigorous overviews. Consequently, this survey does not include studies that were not publicly accessible at the time of the search, such as internal university projects. Additionally, broadly multilingual publications were omitted if they lacked sufficient technical detail or a meaningful language-specific analysis, as this would preclude a proper assessment of their methodology and impact on Tigrinya NLP.

\subsection{Data Extraction and Analysis}
For each included paper, we extracted: authors, year, objectives, methodologies, datasets, key findings, and resources developed. Papers were classified by primary NLP task, but we also allowed multiple tasks per paper when there was substantial treatment of the topic, enabling systematic analysis of progress in each area.

\section{Conclusion}
\label{sec:conclusion}

This work presents a comprehensive analysis of NLP research for Tigrinya, revealing a field characterized by both significant foundational progress and persistent challenges. While resources now exist for several core tasks, substantial gaps remain in data availability, tool development, and advanced applications. Recent years have seen accelerated progress, largely driven by community efforts and targeted research initiatives that leverage cross-lingual methods. The path forward requires combining technical innovation with community engagement to build the culturally relevant, large-scale datasets needed for robust and fair systems. Cross-lingual transfer learning and morphology-aware models offer promising avenues for overcoming data scarcity, while continued investment in language-specific, community-vetted resources remains essential. As NLP technologies increasingly shape global information access, ensuring progress for languages like Tigrinya is not merely a technical challenge but an imperative for linguistic equity and digital inclusion. This survey provides a foundation for researchers, offering an understanding of current capabilities and concrete directions for future work.

\section*{Acknowledgments}
We thank the researchers who reviewed drafts of this work and provided valuable feedback. We invite future contributions in the form of corrections or additions of papers and resources to the open-source \href{https://github.com/fgaim/tigrinya-nlp-anthology}{Tigrinya NLP Anthology} project.

\bibliographystyle{unsrtnat}
\bibliography{references}

\end{document}